\documentclass[lettersize,journal]{IEEEtran}
\usepackage{amsmath,amsfonts}
\usepackage{algorithm}
\usepackage{array}
\usepackage[caption=false,font=normalsize,labelfont=sf,textfont=sf]{subfig}
\usepackage{textcomp}
\usepackage{stfloats}
\usepackage{url}
\usepackage{verbatim}
\usepackage{graphicx}
\usepackage{xcolor}
\usepackage{cite}
\usepackage{arydshln}
\usepackage{algpseudocode}
\usepackage{comment}
\usepackage{booktabs}

\usepackage[pagebackref,breaklinks,colorlinks,allcolors=black]{hyperref}

\usepackage{subcaption}

\begin{document}

\title{RAViT: Resolution-Adaptive Vision Transformer}

\author{Martial Guidez, Stefan Duffner, Christophe Garcia\newline
\small INSA Lyon, CNRS, Université Claude Bernard Lyon 1,\
\small LIRIS, UMR5205, 69621 Villeurbanne, France\
{\tt\small martial.guidez@liris.cnrs.fr, stefan.duffner@liris.cnrs.fr, christophe.garcia@liris.cnrs.fr}
}



\maketitle

\begin{abstract}
Vision transformers have recently made a breakthrough in computer vision showing excellent performance in terms of precision for numerous applications. 
However, their computational cost is very high compared to alternative approaches such as Convolutional Neural Networks.
To address this problem, we propose a novel framework for image classification called RAViT based on a multi-branch network that operates on several copies of the same image with different resolutions to reduce the computational cost while preserving the overall accuracy. 
Furthermore, our framework includes an early exit mechanism that makes our model adaptive and allows to choose the appropriate trade-off between accuracy and computational cost at run-time. 
For example in a two-branch architecture, the original image is first resized to reduce its resolution, then a prediction is performed on it using a first transformer and the resulting prediction is reused together with the original-size image to perform a final prediction on a second transformer with less computation than a classical Vision transformer architecture. 
The early-exit process allows the model to make a final prediction at intermediate branches, saving even more computation. 
We evaluated our approach on CIFAR-10, Tiny ImageNet, and ImageNet. We obtained an equivalent accuracy to the classical Vision transformer model with only around 70\% of FLOPs.
\end{abstract}


\section{Introduction}
\label{sec:intro}

When using computer vision on devices with limited resources like embedded systems, we are interested in reducing the computational cost of a prediction and the size of the hardware, but also the energy consumption, for example to save battery power. 
Recently, the success of transformers~\cite{Vaswani2017AttentionIA} has led to the use of a new type of architecture in computer vision: Vision transformers (ViT)~\cite{Dosovitskiy2020AnII}. 
Since then, ViTs have been successfully applied to several computer vision tasks such as classification~\cite{Dosovitskiy2020AnII,Jiang2021AllTM}, object detection~\cite{Liu2021SwinTH,Wang2021PyramidVT}, and semantic segmentation~\cite{Zheng2020RethinkingSS,Fang2021MSGTransformerEL}.

However, ViTs are computationally expensive due to the self-attention mechanism which scales quadratically with the number of image patches, leading to high memory and computation resource requirements. 
Some research work has been done to try to reduce this cost, mainly by applying methods that have been originally proposed for Convolutional Neural Networks (CNN). 
For example, token pruning~\cite{Li2016PruningFF,Sandler2018MobileNetV2IR} , knowledge distillation~\cite{Hinton2015DistillingTK}, quantization~\cite{Hubara2016BinarizedNN,Krishnamoorthi2018QuantizingDC} and early-exit  architectures~\cite{Demir2024EarlyexitCN,Teerapittayanon2016BranchyNetFI}. 

In addition, there are other approaches that are specific to ViT, sometimes called compact architectures~\cite{Wang2021PyramidVT,Liu2021SwinTH,You2022CastlingViTCS}.
In these compact ViT architectures the computational cost of attention (i.e. the largest part of the total cost) is reduced by changing the way it is computed and thus simplifying the overall computation path. 

Pursuing the same objective of reducing the computational cost, we propose a completely new approach that changes the input resolution to reduce the number of tokens and thus limits the cost of a prediction. 
If an image is reduced to half its size in both dimensions, a ViT would need about 4 times less floating point operations (FLOP) to perform a prediction on it. 
Obviously, reducing the input resolution reduces the accuracy. To address this, our framework performs prediction on several copies of the same image with different resolutions on several computation branches and, in a coarse-to-fine manner, transfers the predictions done by a lower-resolution transformer to the next one operating on a higher-resolution. 
%\autoref{fig:cifar_frame} illustrates this procedure with a two-branch architecture. % 

\begin{figure}[t]
    \centering
    \includegraphics[width=1\linewidth]{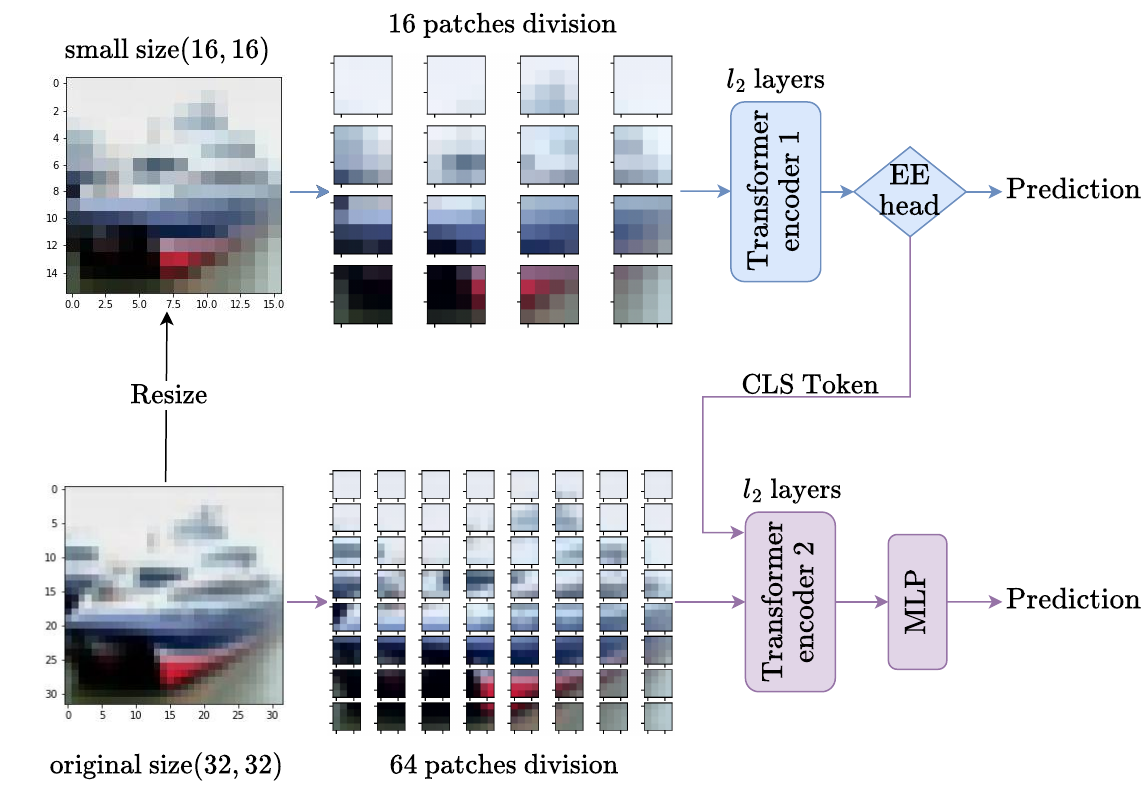}
    \caption{Example of a RAViT architecture on a CIFAR-10 image.}
    \label{fig:cifar_frame}
\end{figure}

Not all images are equally difficult to predict~\cite{Ionescu2016HowHC}, depending on numerous factors related to the quality of the images as well as to their content.
Many dynamic neural network approaches have been proposed in the literature to take advantage of this and save computational resources~\cite{Han2021DynamicNN,Xu2023LGViTDE,Papa2023ASO,Yang2020ResolutionAN,Lin2022SuperVT}.
%S refs
%and more and more optimisations are now dynamic, meaning that a network adapts the prediction cost to the difficulty of the pattern. 
For embedded devices, this is particularly useful and allows the model to predict more with the same energy consumption on average. 
For this reason, we additionally integrated an early-exit (EE) mechanism in our framework:
at inference time, it allows the model to perform fewer computations on simple images and thus enables the overall framework to balance computational cost and accuracy by simply changing the early-exit threshold, which defines the degree of certainty the network requires for a prediction to be considered correct. 
On a low-resource device for example, by adjusting this parameter, we can easily trade off battery over accuracy, or vice versa. % 

We demonstrate through experimentation that our proposed architecture reduces computational costs without sacrificing too much accuracy. We tested our framework on three image classification datasets: CIFAR-10, TinyImageNet, and ImageNet.
For all, a RAViT architecture was found that achieved an accuracy similar to that of a classical ViT, but with only 70\% of the FLOPs required by the original model.

In summary, our method relies on two main contributions: 
\begin{enumerate}
    \item a novel ViT-based multi-branch neural architecture for image classification operating at different resolutions and effectively combining the intermediate predictions in a coarse-to-fine way,
    \item an early-exit mechanism allowing to dynamically control the balance between computational cost and accuracy at run time.
\end{enumerate}

\section{Related Work}
\label{sec:related}

This section provides a brief overview of the background to Transformers, Vision Transformers and the compression methods used on these architectures.\medskip

%--------------------------------------------------------------------------
\noindent\textbf{Transformer.} The transformer architecture was first introduced by Vaswani et al.~\cite{Vaswani2017AttentionIA} for the task of machine translation. 
It relies on self attention combined with feed forward layers and is able to learn strong long-range dependencies in sequences.
They have then quickly been adopted for other Natural Language Processing (NLP) tasks providing state-of-the-art performance on several benchmarks~\cite{Devlin2019BERTPO, Radford2018ImprovingLU}. 
The core component in transformer models is a self-attention mechanism that enables parallel processing and effective handling of long-range dependencies. This allows transformers to capture complex relationships between tokens across an entire sequence, making them highly efficient and scalable.\medskip

%--------------------------------------------------------------------------
\noindent\textbf{Vision Transformer (ViT).} Traditional CNNs have long dominated the field of computer vision, primarily due to their ability to capture spatial hierarchies through convolutional layers compared to classic non-convolutional Deep Neural Networks (DNN). 
Inspired by the success of transformers in NLP, Dosotovitsky et al. \cite{Dosovitskiy2020AnII} have developed ViT for image recognition. 
In contrast to CNN, ViT performs image analysis by splitting images into fixed-size patches (equivalent to a token in a traditional transformer). 
These image patches are treated as tokens and processed using a transformer. 
Compared to CNN, ViT captures global image context more effectively especially where long-range dependencies are crucial. However, it usually needs to be trained on very large-scale datasets like JFT-300M~\cite{Sun2017RevisitingUE}. 
This requirement comes from the lack of inductive biases inherent in CNN models \cite{Dosovitskiy2020AnII}. 
Thus, to use them on smaller datasets, it is better to use models that have been pre-trained on a large dataset.
The ViT design has inspired a wider exploration of transformer-based architectures for vision tasks, with applications beyond image classification spanning object detection~\cite{Carion2020EndtoEndOD}, segmentation~\cite{Thisanke2023SemanticSU}, and even video understanding~\cite{Arnab2021ViViTAV}.\medskip

%--------------------------------------------------------------------------
\noindent\textbf{ViT Compression.} There are two types of compression methods: static and dynamic. Static methods focus on reducing network complexity regardless of the input. On the other hand, dynamic methods~\cite{Han2021DynamicNN} produce models that adapt themselves at execution time according to the input.

Many compression methods proposed for CNN have been applied to ViT~\cite{Papa2023ASO}. 
One example is token pruning~\cite{Zhu2021VisionTP}. This method aims at suppressing some tokens that are not useful for the classification in order to perform the computation on only a fraction of the total number of tokens. 
Knowledge distillation has also been applied to ViT~\cite{Wu2022TinyViTFP}, where a "student" network is trained by imitating a larger pre-trained network called "teacher". 
The goal is to get a higher level of accuracy of the student network than if it was trained from scratch. 
Quantization~\cite{Liu2022NoisyQuantNB} aims to reduce the size of the network and the computational cost by reducing the encoding precision of weights or activations (from floats32 to int8 for example). 
Other approaches rely on so-called early exits~\cite{Xu2023LGViTDE}. 
Here the idea is to split a network into several blocks or branches of different complexity with associated side exits. If the network is sure of the prediction at the end of a branch, it will exit earlier without doing the rest of the computation. 

Some compression methods are specifically designed for ViT, such as compact architectures~\cite{Wang2021PyramidVT,Liu2021SwinTH,You2022CastlingViTCS}, which aim to change the design of the attention block to reduce its computational cost. 
In fact, this can be seen as the computational bottleneck as its complexity increases quadratically with the number of patches within the image 
due to two dot product operations in self-attention. \medskip

Closer to our work, some existing methods propose some sort of multi-resolution or gradual refinement strategies for ViT or CNN models.
For example, Super Vision Transformer~\cite{Lin2022SuperVT} uses token pruning combined with multi-scale patching to reduce the computational cost of the network. This method changes the patch size applied on the image in the different branches, and then applies pruning to each patched copy of the image to remove the uninformative tokens. 
CF-ViT~\cite{Chen2022CFViTAG} aims to refine some tokens that are considered informative to perform more specific calculations on those areas. 
The image is first patched and a prediction is performed. If the prediction is not certain enough, the network identifies the tokens that are the most informative and then refines these tokens.
Finally, some CNN-based methods vary the input resolution~\cite{Yang2020ResolutionAN}, the network starts with a low-resolution version of the input and if the image is hard to classify, the resolution is progressively increased until the original resolution. 
This approach is quite similar to ours. However, due to the use of Transformer instead of CNN, the feature transfer process is very different and does not depend on each branch architecture. We only transfer the classification (CLS) token (detailed in the next section) without any need of transferring features to blocks individually and designing specific transfer layers (like in RANet~\cite{Yang2020ResolutionAN}).
There are also non-dynamic methods such as Multiscale Vision Transformer (MViT)~\cite{Fan2021MultiscaleVT}, which is a multi-channel resolution scale framework. This method uses a special attention architecture called Multi Head Pooling Attention and a reduction of the spatio-temporal resolution (i.e. sequence length) without reducing the input resolution.\smallskip

\section{Method}
\label{sec:method}

Our proposed approach consists in a principled framework to efficiently and dynamically extract information at different resolutions of the input, from coarse to fine, where the coarser features contribute to the processing of the finer ones and where the latter are only used if necessary.\smallskip

\begin{figure*}[ht]
    \centering
    \includegraphics[width=0.9\linewidth]{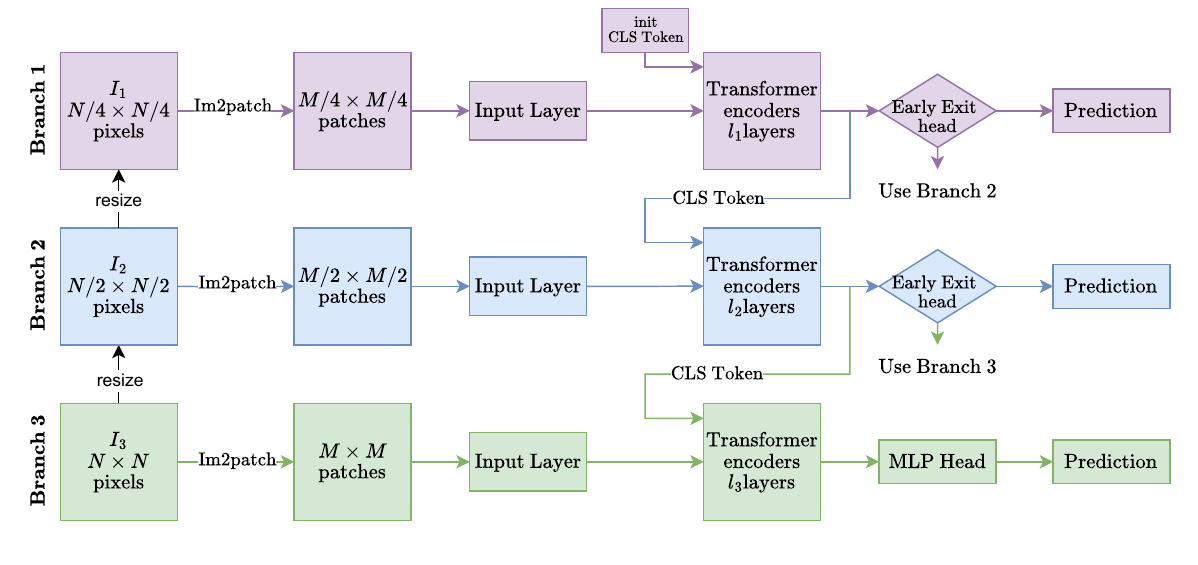}
    \caption{Framework of our RAViT architecture with 3 branches. Note that the number of branches can be increased or decreased without changing radically the architecture}
    \label{fig:fram}
\end{figure*}

\textbf{Inference.} 
The overall inference procedure for a RAViT architecture trained with $B$ branches is as follows: a given input image $I$ ($N \times N$ pixels) is scaled to produce $B$ images (where $I_B$ is the original image without size reduction), $I_1, \ldots, I_B$, each at a different resolution, $\texttt{dim}(1), \ldots, \texttt{dim}(B)$. The processing begins with the lowest-resolution image, $I_1$.
It will be passed to the first ViT encoder $T_1$ having $l_1$ transformer layers and constituting the first branch of the architecture. 

Subsequently, an early-exit head will execute a primary prediction on the classification token and calculate the uncertainty score to determine the adequacy of the prediction. The head composition is characterized by its parsimonious nature, representing the aggregate of an MLP layer and the execution of the uncertainty calculation (using the entropy of the prediction softmax).
In the event that the computed uncertainty exceeds the established threshold ($E_{th}$), an additional prediction is conducted on $I_2$ with $T_2$ (equipped with $l_2$ transformer layers) on the second branch. This process utilizes the classification token outputted by the first branch as the second branch's initial classification token, thereby avoiding a complete reinitialization. Subsequently, an evaluation of its uncertainty is performed.

The process is repeated iteratively until the confidence level falls below the established threshold. In instances where the confidence level is inadequate prior to the final branch, the final prediction will be determined by it.\smallskip

\textbf{Information transmission through branches.}
For all transformer encoders $T_i$ the patch size remains the same, and each transformer adapts the number of patches ($M^2$ for the original image) to the input size. 
This allows us to use the same embedding size and hidden dimension for all transformers. 
To avoid losing extracted information between the different transformers, we need a way of transferring these higher-level features from one encoder to the next. 
There are many possible solutions, and we chose a straightforward but effective one: 
we simply retrieve the classification (CLS) token produced by a transformer encoder, $cls_i$ and give it as input to the next encoder $T_{i+1}$. As explained earlier, all transformers have the same dimensions, which ensures that the dimension of the CLS token remains the same on all branches.

\autoref{alg:Ravit} summarizes the overall inference algorithm.\smallskip

\begin{algorithm}[t]
    \small
    \begin{algorithmic}
    %\SetKwInOut{Input}{input}
    %\SetKwInOut{Output}{output}
    %\Input{Image I}
    %\Output{Classification}
    \State $i \gets 1$
    \State $cls_{i-1} \gets $rand() \Comment{Initialize CLS with random vector}
    \While{not $\mathrm{conf}$ or $i \leq B$}
    \State $I_i \gets$ resize($I$, dim($i$)) \Comment{Resize image}
    \State $cls_{i} \gets$ ViT($I_i$, $l_i$, $cls_{i-1}$) \Comment{Propagate through $T_i$}
    \State $p \gets $MLP$_i(cls_{i})$ \Comment{Compute prediction}
    \State $c \gets$ entropy(softmax($p$)) \Comment{Compute uncertainty}
    \State $i ++$
    \State $\mathrm{conf} \gets c<E_{th}$ \Comment{Threshold uncertainty}
    \EndWhile
    \State \Return arg max($p$)
    \end{algorithmic}
 \caption{RAViT inference algorithm.}
 \label{alg:Ravit}
\end{algorithm}

It is evident that a 2-branch (see \autoref{fig:cifar_frame}) or 4-branch architecture would function in a similar manner. 
It is also noteworthy that a resize factor of 2 was employed for the diverse input image dimensions, denoted by $\texttt{dim}(i)$, across each branch, $i$. However, it is also conceivable to employ a finer scale gradation, such as 1.2 or 1.5, as an alternative. \smallskip

\textbf{Early-exit.} Our early exit method is based on the BranchyNet architecture~\cite{Teerapittayanon2016BranchyNetFI}, the decision process is based on the entropy of the classification result. After calculating the softmax of the output ($p$ in \autoref{alg:Ravit}), the confidence score is defined as the entropy of this softmax vector ($c$ in \autoref{alg:Ravit}). Then, we define a threshold ($E_{th}$) that decides if the sample classification is confident enough. 
The lower the threshold, the more reliable the result. For each early exit, we can adjust the threshold individually. \smallskip

\textbf{Training Loss.} For the training process, we minimize a global loss, which is the sum of each exit loss weighted by a chosen coefficient $\omega_i$

\begin{equation}
    \mathcal{L}_{total} = \sum_{i=1}^{B} \omega_i\mathcal{L}_{branch\text{-}i},
    \label{eq:loss}
\end{equation}

(cf.\  \autoref{eq:loss} for a RAViT with $B$ branches). For each exit, the loss function is cross entropy. The choice of these coefficients is discussed in the next section.\smallskip

\textbf{Computational Cost.}\label{sec:comput} The computational cost of a transformer layer of the $i$th branch is given by:

\begin{equation}
    \begin{split}
        \text{MAC}_{t_i} = 2L_i^2D_{emb} + 12L_iD_{emb}^2\ .
    \end{split}
    \label{eq:macs}
\end{equation}

Where $L_i$ is the sequence length of the $i$th branch (we add one to the patch number to account for the CLS token), $D_{emb}$ is the channel size of a token (embedded dimension). MAC means Multiply-ACcumulate operation. In this formula, the first term correspond to the attention computation and the second to the linear projections and MLP computation. As the computational complexity of patch embedding can be neglected compared with the transformer layer, we will use \autoref{eq:macs} as the total computation cost of our models. \smallskip

The total number of MACs is given by:

\begin{equation}
    \text{MAC}_{tot} = \sum_{i=1}^{B} l_i\times\text{MAC}_{t_i}\ ;
    \label{eq:macs_tot}
\end{equation}

Where $l_i$ correspond to the number of layers on the $i$-th transformer. By definition, the number of MACs is the half of the number of Floating Point Operations (FLOPs). For the experiments, we will use $\text{FLOPs} = 2\times\text{MAC}_{tot}$.\smallskip

One of the most well-known disadvantages of early exit is the loss of computation. It is considered an inefficient use of resources if the exit head of the different branches performs computations that are not subsequently utilized. However, within the framework proposed, the computational cost associated with these exit branches is minimal. The framework consists of a single fully-connected linear layer and entropy calculation.\smallskip

\begin{figure*}
    \centering
  \subfloat[\label{fig:map_cifar_acc}]{%
       \includegraphics[width=.49\linewidth]{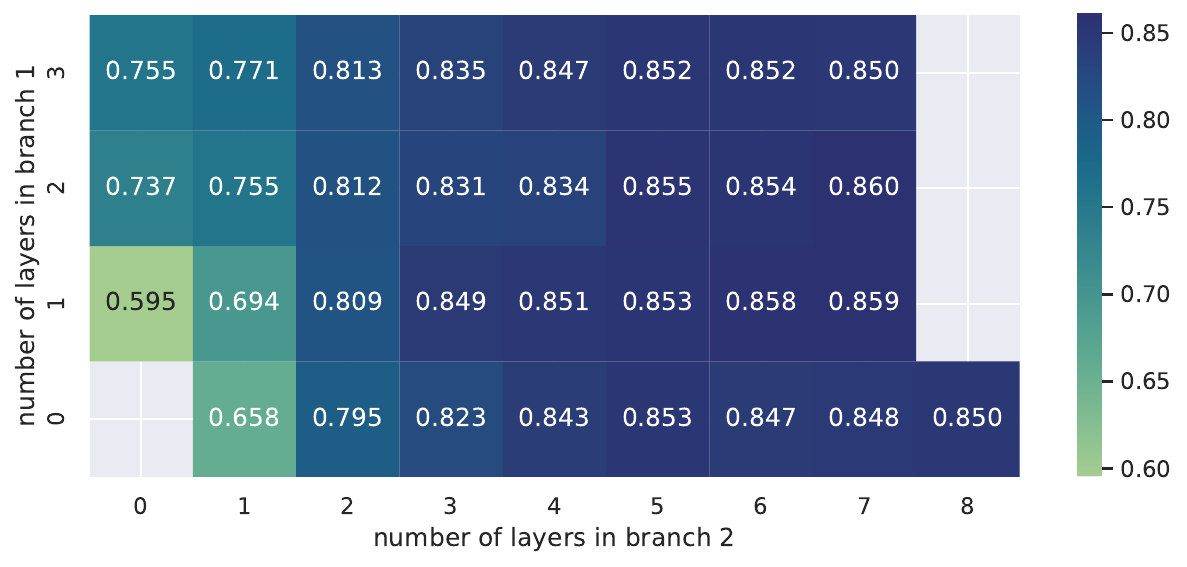}}
  \subfloat[\label{fig:map_cifar_flops}]{%
        \includegraphics[width=.49\linewidth]{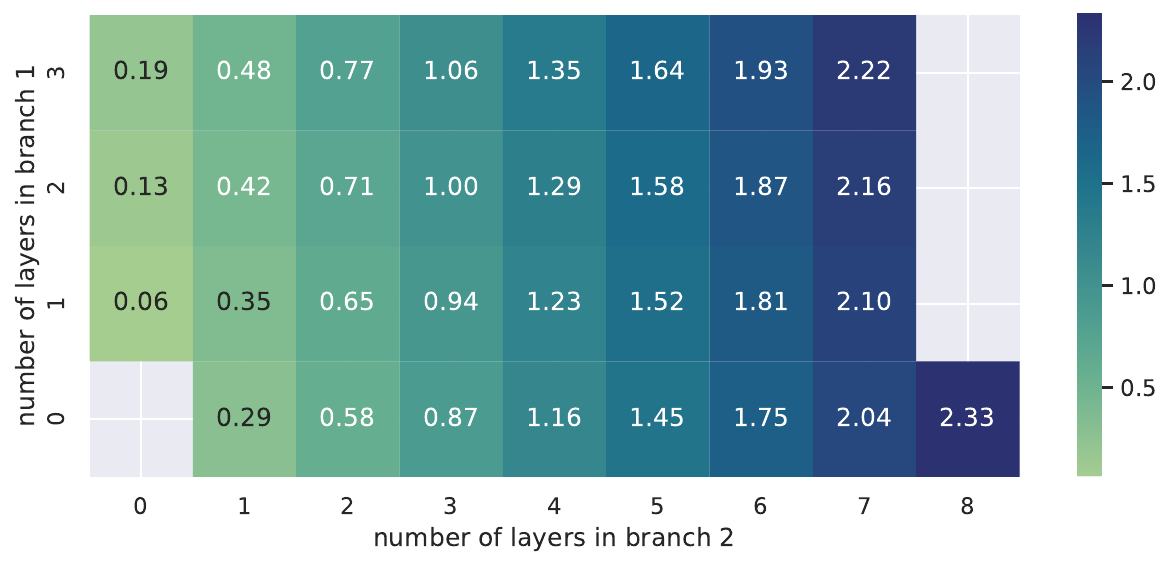}}
    
    \caption{Performance analysis of different RAViT models with 2 branches used on CIFAR-10. (a) Accuracy for different layer numbers on CIFAR-10. (b) Number of FLOPs for different layer numbers on CIFAR-10.}
    \label{fig:heatmap_cifar}
\end{figure*}

To determine the computational cost with the early exit ($\text{FLOP}_{EE}$), as it depends on which exit is used, we averaged it using the number of samples exited at each branch multiplied by the cost of the network up to the exit. Note that $S_{tot}$ is the number of images we use to calculate the accuracy, $S_i$ is the number of images that exit at the $i$th exit, and $\text{FLOP}_i$ is the computation cost for an image that exits at the $i$th branch.\smallskip 

\begin{equation}
    \text{FLOP}_{EE} = \frac{\sum_{i=1}^{B} S_i \times \text{FLOP}_i}{S_{tot}}
\end{equation}

\section{Experiments}
\label{sec:experiments}

We experimentally evaluated our method on three datasets with different image sizes: CIFAR-10~\cite{Krizhevsky2009LearningML}, Tiny ImageNet~\cite{Le2015TinyIV}, and ImageNet \cite{deng2009imagenet}.
Note that we focus our evaluation on various variants of RAViT with different numbers of branches and different number of layers in each branch showing the flexibility of our method, and a clear reduction in computation. Thus, our aim was to provide a proof of concept rather then to optimise the classification accuracy for the different benchmark datasets which would have required to first pre-train our models on a large dataset like ImageNet or JFT-300M. This is quite resource-demanding and may also induce some hidden bias in the study. As a consequence, we did not do a comparison to other ViT compression methods. But we do not see them as competitors, and as our approach is completely different from them, it could be combined with most of them to obtain even larger computational reductions. 

% CIFAR-10
The images constituting the CIFAR-10 dataset are of dimensions $32\times32$ pixels. 
However, dividing this size by four would be excessive. Therefore, for these images, a 2-branch architecture is employed, wherein the image is resized to $16\times16$pixels for the first branch, and the other branch utilizes the original image. These images have been randomly cropped to  the same size with a padding of 4, also randomly horizontally flipped and normalized using:$\texttt{mean}=[0.4914, 0.4822, 0.4465]$, $\texttt{std}=[0.2023, 0.1994, 0.2010]$

The training hyper-parameters are based on existing literature, e.g. Kentaro Yoshioka~\cite{yoshioka2024visiontransformers}. Following the execution of several tests on a classic 6-layer transformer architecture, it was determined that the subsequent parameters would be employed for all subsequent tests: a batch size of 256, 400 training epochs, we use AdamW~\cite{Loshchilov2017DecoupledWD} optimiser (with a learning rate set to $\texttt{lr}=10^{-3}$,exponential decay $\beta_1 = 0.9$, $\beta_2 = 0.999$, and a weight decay $\texttt{wd}=0.1$), CosineAnnealingLR ~\cite{Loshchilov2016SGDRSG} as scheduler, a patch size of (4,4) pixels, 128 embedded dimension, and 512 hidden dimension.

Our first experiment evaluates the 2-branch architecture with
different numbers of layers in each branch ($l_1 \in [\![0;3]\!]$ and $l_2 \in [\![0;7]\!]$). 
The idea is to perform a rather exhaustive study showing the performance of all possible variants and then to highlight some models with specific combinations of $l_1$ and $l_2$ that give a comparable or better accuracy than the baseline but with a lower number of FLOPs.

As explained in the previous part, the use of early-exit induce a need to set specific loss coefficients. Having this coefficient set to $1/2$ for 2 branch architecture gives us good results (comparing to other tested arbitrary choices). We also set the batch size to 1 when using the network after the training phase, as in~\cite{Teerapittayanon2016BranchyNetFI}. This is more practical in the case of early exit in order to know the exit used for each sample. Otherwise, each sample of a batch needs to be routed individually to different exits, which may induce an additional computational cost. Moreover, as mentioned earlier, one appropriate target application scenario of our approach is in the context of real-time embedded systems which would process a stream of images sequentially.

% Tiny ImageNet
Tiny ImageNet contain $64\times64$ down-sampled images from ImageNet. 
There are 200 classes and for each, there are 500 training images, 50 validation images and 50 test images. The images are big enough to use 3 branch architecture. For the 3-branch architectures, we use loss coefficient set to $1/3$. The images are randomly cropped resized to $64\times64$ pixels and randomly horizontally flipped before normalizing them with: $\texttt{mean}=[0.485, 0.456, 0.406]$, $\texttt{std}=[0.229, 0.224, 0.225]$. Our models have been trained for 50 epochs using AdamW optimizer with a learning rate of $\texttt{lr} = 10{-4}$, the exponential decays are set to $\beta_1 = 0.9$ and $\beta_2 = 0.99$ and the weight decay has a value of $\texttt{wd} = 0.1$. Cosine annealing scheduling has been applied to the learning rate.

For ImageNet, we use training parameters inspired by Alexey Dosovitskiy work \cite{Dosovitskiy2020AnII}. We choose to use AdamW optimizer with a learning rate of $\texttt{lr} = 3\time10{-3}$, the exponential decays are set to $\beta_1 = 0.9$ and $\beta_2 = 0.99$ and the weight decay has a value of $\texttt{wd} = 0.3$. Cosine annealing scheduling has been applied to the learning rate.
In this dataset, the images are randomly crop resized to 224*224 pixels, randomly horizontally flipped and normalized using the following parameters: $\texttt{mean}=[0.485, 0.456, 0.406]$, $\texttt{std}=[0.229, 0.224, 0.225]$. Since TinyImageNet is a resized version of ImageNet, this explain the use of the same normalization parameters on the two datasets. Similarly to its Tiny version, for this dataset, we use a 3 branch RAViT with 16*16 pixels patches. The images are resized to 64*64 pixels in the branch 1 and 128*128 pixels in branch 2. Our models have been trained for 500 epochs approximately.

\section{Results}
\label{sec:results}

\begin{figure*} 
    \centering
  \subfloat[\label{fig:tiny_global}]{%
       \includegraphics[width=.47\linewidth]{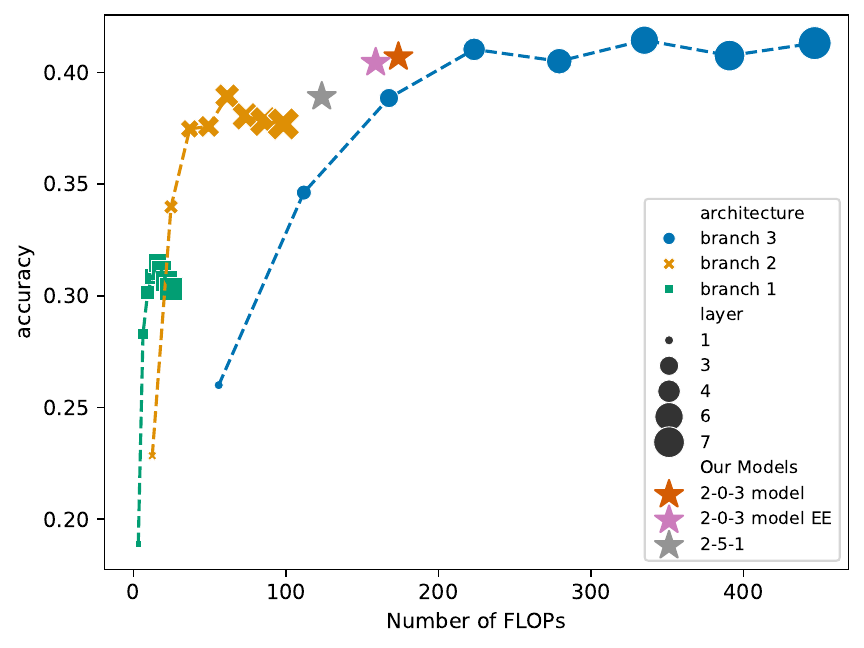}}
    \hfill
  \subfloat[\label{fig:2-0-3tiny}]{%
        \includegraphics[width=.49\linewidth]{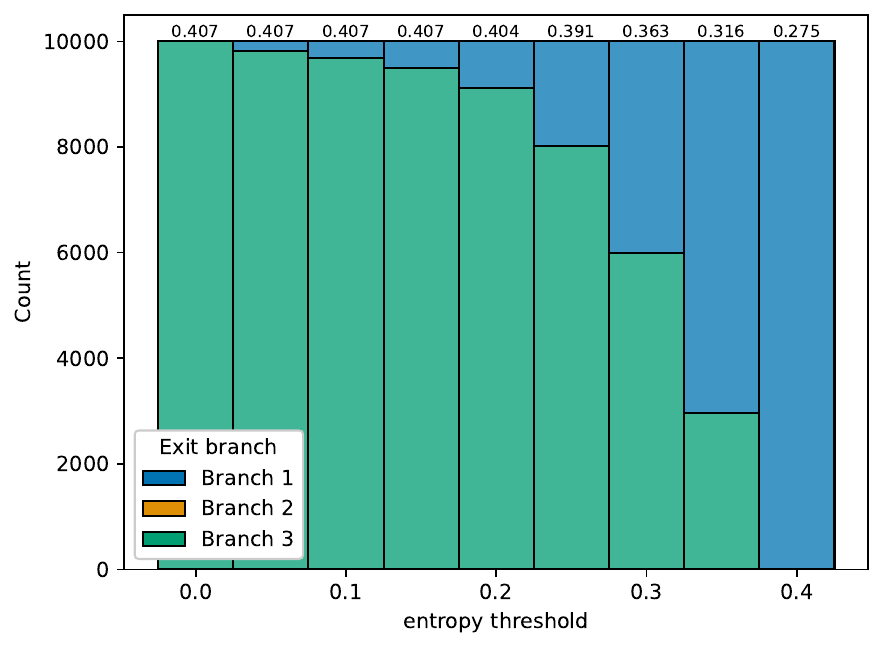}}
  \caption{Tiny ImageNet results. (a) Test accuracy in relation to FLOPs for different architectures. (b) Accuracy (written on top of each columns) and exit distributions for different values of the early-exit threshold applied to a 2-0-3 network.}
  \label{fig1} 
\end{figure*}

In this part, an $i$-$j$ model will refer to a model with $i$ layers in the first branch and $j$ layers in the second one. 
Similarly, an $i$-$j$-$k$ model will refer to a model with $i$ layers in branch-1, $j$ layers in branch-2 and $k$ layers in branch-3. 
Single type of branch networks refer to networks that use only one of their branches. 
For example a 0-0-$k$ network is a classic $k$-layer ViT and a 0-$j$-0 network is an $j$-layer ViT using half-size images. 
"No EE" refer to a model that have a early exit threshold set to 0.

It must be acknowledged that, in view of the non-normalization of the EE-threshold, the range of this parameter fluctuates across disparate datasets.

\subsection{CIFAR-10}

(\autoref{fig:map_cifar_acc}) shows the test accuracy for different architectures with different numbers of layers on each branch. 
Thus each value corresponds to a different architecture with two branches. 
(\autoref{fig:map_cifar_flops}) shows the number of GFLOPs for each of these architectures. 
These results show that adding a layer on the first branch to an architecture with a certain number of layers on the second branch increases the accuracy. 
However, it is important to note that increasing the number of branch-1 layers is not always beneficial in terms of accuracy, for example the 1-3 model obtained better accuracy than the one with 2 or 3 branch-1 layers and 3 branch-2 layers. % 

\begin{table}[ht]
    \centering
    \vskip 0pt
    \caption{Comparison of ViT with different number of layers and our models with several EE thresholds on Cifar-10.}
    \label{tab:CIFAR}
    \small
    \begin{tabular}{@{}l@{\hskip 0pt}c@{\hskip 4pt}c@{\hskip 4pt}c@{}}
        \toprule
        Architecture & Accuracy & $\ $MFLOPs & $\ $\% MFLOPs\\
        \midrule
        3 layers ViT & 82.3 & 83.16 & 75 \\
        4 layers ViT & 84.3 & 110.88 & 100 \\
        5 layers ViT & 85.3 & 138.61 & 125 \\ \midrule
        our 1-3 (no EE) & 84.9 & 89.99 & 81 \\
        our 1-3 (0.05 $E_{th}$) & 84.6 & 80.81 & 72 \\
        our 1-3 (0.15 $E_{th}$) & 82.6 & 67.72 & 61 \\
        \bottomrule
    \end{tabular}
\end{table}

We can see that the 1-3 model is interesting because it has a better accuracy than the 4 branch-2 layers model and is just below the accuracy of the 5 branch-2 layers model. The computational cost of this model is about $0.94$ GFLOPs, which is a reduction of $19\%$ compared to the 4 layers model and $35\%$ compared to the model with 5 layers on branch-2. % 

It is also interesting to note that increasing the number of branch-2 layers increases the accuracy up to a certain point (here 5 branch-2 layers), and adding more layers on the second branch does not increase the accuracy. 
But adding layers on the first branch may still improve the model slightly, e.g. we get an accuracy of $86.0\%$ for a 2-7 model. % 

The previous results are without considering the early exits. It will be discussed on the TinyImageNet results.

\begin{comment}
Let us now look at the effect of the early-exit entropy threshold on the 1-3 model. 
\autoref{fig:1-3cifar} shows the proportion of images that have an exit on each branch (here, two) for different entropy thresholds and the associated accuracy. 
Intuitively, increasing the threshold increases the proportion of images that exit on the first branch. 
Above 0.4, all images exit on branch-1. 
It is also interesting to note that when the threshold is set to 0.05, the accuracy decreases by $0.3\%$ and the computational cost decreases by $7\times10^{-3}$ GFLOPs, i.e. a $10\%$ reduction. %  
\end{comment}

\autoref{tab:CIFAR} summarises the results obtained on CIFAR-10 comparing some relevant original ViT with a RAViT with and without early exit for a given early exit threshold. 
Overall, our method achieves a computational reduction of $28\%$ maintaining the accuracy and $44\%$ with a decrease of only $1.7\%$ points in accuracy.

\begin{figure*} 
    \centering
  \subfloat[\label{fig:map_tiny_acc}]{%
       \includegraphics[width=.45\linewidth]{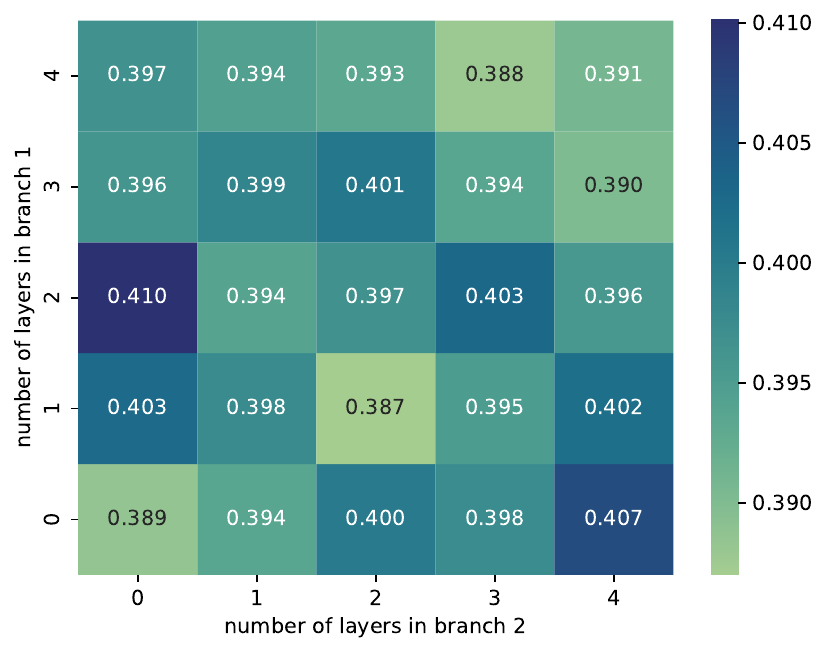}}
    \hfill
  \subfloat[\label{fig:map_tiny_flops}]{%
        \includegraphics[width=.45\linewidth]{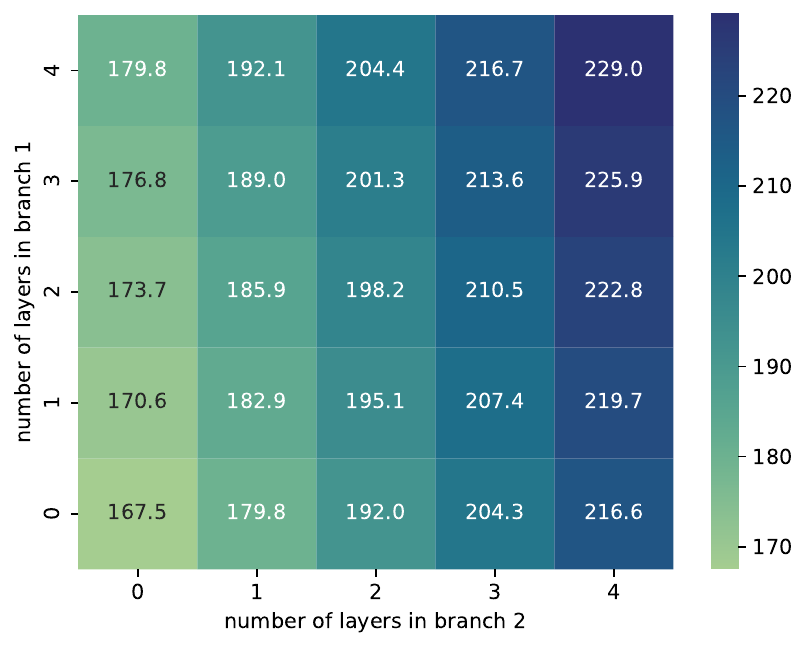}}
    
  \caption{Performance analysis of different RAViT trained on Tiny ImageNet with 3 branches and 3 layers on the third one. To be compared with a 4-layer classical ViT of 0.41 accuracy and 223.3 GFLOPs. (a) Accuracy for different models having 3 branch-3 layers. (b) GFLOPs for different models having 3 branch-3 layers.}
  \label{fig:heatmap_tiny}
\end{figure*}

\subsection{Tiny ImageNet}

We performed similar tests on Tiny ImageNet with a 3-branch architecture (cf.\ \autoref{fig:fram}). 
First, we compared the various single-branch architectures.
\autoref{fig:tiny_global} shows the accuracy of these different models with the corresponding number of FLOPs. 
As mentioned in \autoref{sec:comput}, there is about an order of 4 between the cost of calculating a branch-2 and branch-1 layer and a branch-3 and branch-2 layer. 
We can also note that a 5 branch-2 layers network has a better accuracy then a 3 branch-3 layers network (which is 2.73 times more computationally expensive). 
This first result shows that sometimes, it can be interesting to simply reduce the input size to reduce the computational cost of a network.
Also, at some point, adding more layers does not increase the accuracy anymore.

After having studied the single-branch architectures, we evaluated several architectures with a various number of layers in each branch. 
A 2D visualisation (similar to \autoref{fig:heatmap_cifar}) is difficult because of the third dimension here (one for each branch). 
Instead, we will look at maps where the number of branch-3 layers is fixed and the vertical and horizontal axes show the number of branch-1 and branch-2 layers. 
For example, \autoref{fig:heatmap_tiny} shows the results obtained for 3 layers in branch-3.
We can see that the 2-0-3 model is very interesting in terms of accuracy compared to computational cost. 
On \autoref{fig:tiny_global}, we added a point that corresponds to this model to compare it with the reference models. 

\begin{table}[ht]
    \vskip 0pt
    \centering
    \caption{Comparison of ViT with different number of layers and our models with several EE thresholds on Tiny ImageNet.}
    \label{tab:Tiny}
    \small
    \begin{tabular}{@{}l@{\hskip 0pt}c@{\hskip 5pt}c@{\hskip 8pt}c@{}}
        \toprule
        Architecture & Accuracy & $\ $GFLOPs & \% GFLOPs\\
        \midrule
        3 layers ViT & 38.8 & 11.52 & 75 \\
        4 layers ViT & 41.0 & 15.36 & 100 \\
        6 layers ViT & 41.4 & 23.04 & 150 \\ \midrule
        our 2-0-3 (no EE) & 40.7 & 12.00 & 78 \\
        our 2-0-3 (0.2 $E_{th}$) & 40.4 & 10.98 & 71 \\
        our 2-0-3 (0.25 $E_{th}$) & 39.1 & 9.72 & 63 \\
        \bottomrule
    \end{tabular}
\end{table}

For the 2-0-3 model, we have made a study of the impact of the entropy threshold. The results are illustrated in \autoref{fig:2-0-3tiny}.
The accuracy is maintained up to an entropy threshold of 0.15, with a computational reduction of 5\% compared to the model without early exit. 
With a threshold of 0.2, we lose $0.03\%$ of accuracy with a computational reduction of 8.5\%.

\autoref{tab:Tiny} shows how the 2-0-3 model performs (without early exit and with a early exit threshold of 0.2) on Tiny ImageNet compared to classical ViT with 3 or 4 layers. % 
We obtained a computation reduction of 29\% with a negligible decrease in accuracy (0.6\% points) and a 37\% reduction with only 1.9\% points lower accuracy.

\begin{figure*} 
    \centering
  \subfloat[\label{fig:imagenet_1-1-8}]{%
       \includegraphics[width=.49\linewidth]{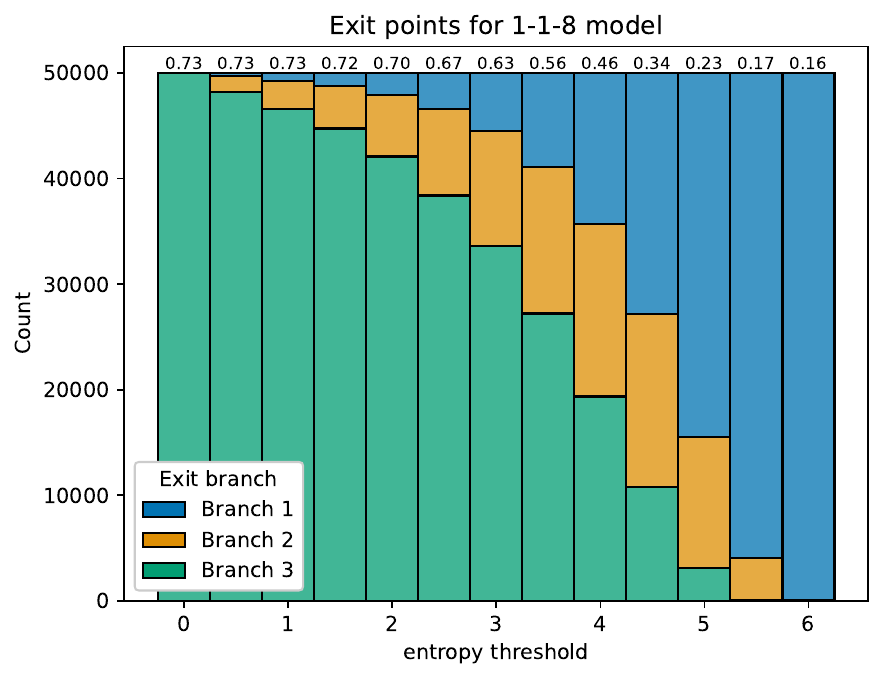}}
    \hfill
  \subfloat[\label{fig:imagenet_1-1-10}]{%
        \includegraphics[width=.49\linewidth]{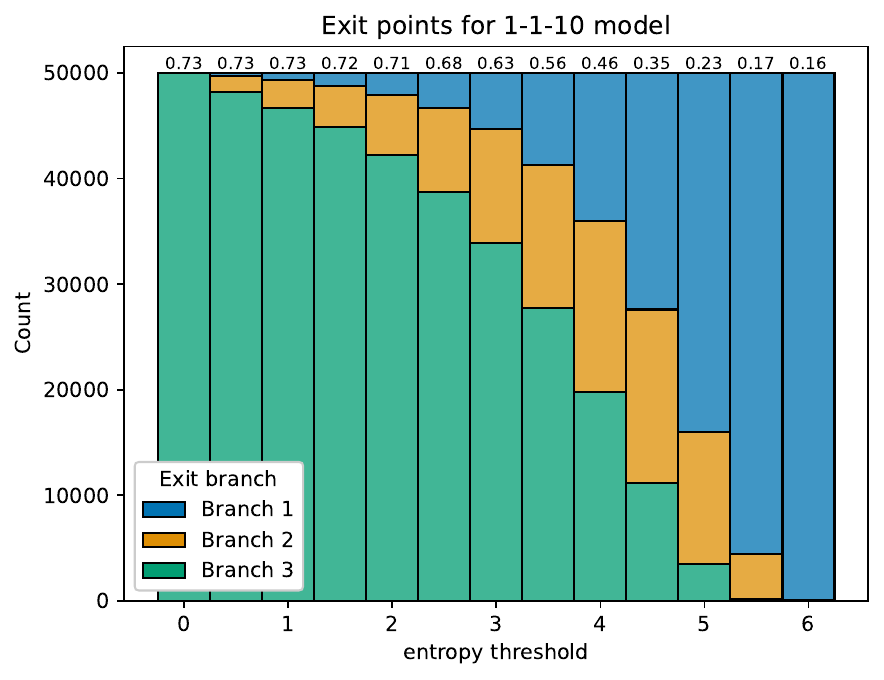}}
  \caption{Accuracy (written on top of each columns) and exit distributions for different values of the early-exit threshold applied to different models on ImageNet. (a) RAViT 1-1-8 model. (b) RAViT 1-1-10 model.}
  \label{ImageNet_result} 
\end{figure*}

\subsection{ImageNet}

A series of RAViT models were examined, with ViT-B (a 12-layer ViT) and also 8 and 10 layer ViT serving as reference. All models have been trained from scratch and due to the high training cost, we were not able to test a lot of hyper parameters and data augmentation. 
We also need to note that the accuracy of our models are not as high as the state of the
art. We were not able to tune the training parameters optimally. Nevertheless, we claim that it does not change the quality of this results. It is also the main obstacle for us to compare our method with other state of the art compression techniques that either used pretrained models or have been trained from scratch and achieve to reach the a Top 1 accuracy of ViT-B model.

\begin{table}[ht]
    \vskip 0pt
    \centering
    \caption{Comparison of ViT with different number of layers and our models on ImageNet.\label{ravit_tab}}
    \label{tab:imagenet}
    \small
    \begin{tabular}{@{}l@{\hskip 0pt}c@{\hskip 5pt}c@{\hskip 8pt}c@{}}
        \toprule
        Architecture & Accuracy & $\ $GFLOPs & \% GFLOPs\\
        \midrule
        8 layers ViT & 71.08 & 23.26 & 67 \\ 
        10 layers ViT & 72.80 & 29.07 & 83 \\
        12 layers ViT (ViT-B) & 73.36 & 34.89 & 100 \\ \midrule
        our 1-1-8 (no EE) & 73.25 & 24.43 & 70 \\
        our 1-1-8 (1.0 $E_{th}$) & 72.6 & 22.84 & 65 \\
        our 1-1-10 (no EE) & 73.38 & 30.25 & 81 \\
        our 1-1-10 (1.5 $E_{th}$) & 72.3 & 27.23 & 78 \\
        \bottomrule
    \end{tabular}
\end{table}

The results presented in \autoref{fig:imagenet_1-1-8}, \autoref{fig:imagenet_1-1-10}, and \autoref{ravit_tab} demonstrate that our method yields analogous outcomes for this particular dataset as previously observed with the two other. The 1-1-8 RAViT model demonstrated a relative 99.85\% accuracy rate when compared to the ViT-B reference model, with only 70\% of the computation cost.

The study of the EE demonstrated analogous results to those observed in the other datasets. As the entropy threshold increases, there is a corresponding decrease in both accuracy and computational complexity.

\section{Discussion}
\label{sec:discussion}

The most critical parameters are those that define the neural architecture, i.e., the number of branches and layers in each branch. 
To date, a straightforward formula for determining the ideal number of layers on each branch remains unidentified. 
One potential solution to this problem would be to apply a Neural Architecture Search (NAS) algorithm to our framework in order to automate the process. 

A further avenue of research would involve correlating the early-exit threshold with a hardware parameter and examining the model's behavior under the dynamics of a data stream of varying complexity. 
In the context of an embedded system deployment, the EE threshold can be augmented when the battery percentage diminishes. This adjustment is intended to conserve power and extend the battery's operational longevity. % 

It is also possible to modify the EE thresholds for each branch independently or to make changes to the branch architecture. 
In the context of an architectural framework characterized by three distinct resolutions, the initial branch may encompass the layers applied to the coarse image, in addition to the intermediate layers (either fully or partially). Furthermore, it is possible to subdivide the layers applied to the original image into two distinct branches.

\section{Conclusion}
\label{sec:conclusion}

We have proposed RAViT, a novel ViT-based framework for image classification that performs predictions on different image resolutions. 

The proposed method yielded satisfactory results on three distinctly different image classification datasets, employing an approach that is relatively uncomplicated and does not necessitate extensive hyper-parameter optimization. It has been demonstrated that the accuracy obtained is comparable to that of the classic network, while the computational cost is only 70\% of that of the classic network.

RAViT is an architecture optimization for ViT that may be especially interesting for a deployment on embedded devices to have the possibility to dynamically adapt the network balance between preserving the computation cost and having precise predictions.

\section{Acknowledgement}

This work was granted access to the HPC resources of IDRIS under the allocation 2024-25 AD011015765 and AD011015765R1  made by GENCI. 

The authors acknowledge the ANR – FRANCE (French National Research Agency) for its financial support of the RADYAL project n°23-IAS3-0002.

\bibliographystyle{IEEEtran}
\bibliography{main}

@String(CVPR= {IEEE Conf. Comput. Vis. Pattern Recog.})

@String(ICCV= {Int. Conf. Comput. Vis.})

@String(ICPR = {Int. Conf. Pattern Recog.})

@String(ICLR = {Int. Conf. Learn. Represent.})

@String(AAAI = {AAAI})

@String(CVPR  = {CVPR})

@String(ICCV  = {ICCV})

@String(ICPR  = {ICPR})

@String(ICLR  = {ICLR})

@inproceedings{Vaswani2017AttentionIA,
 author = {Ashish Vaswani and
Noam Shazeer and
Niki Parmar and
Jakob Uszkoreit and
Llion Jones and
Aidan N. Gomez and
Lukasz Kaiser and
Illia Polosukhin},
 bibsource = {dblp computer science bibliography, https://dblp.org},
 booktitle = {Advances in Neural Information Processing Systems 30: Annual Conference
on Neural Information Processing Systems},
 editor = {Isabelle Guyon and
Ulrike von Luxburg and
Samy Bengio and
Hanna M. Wallach and
Rob Fergus and
S. V. N. Vishwanathan and
Roman Garnett},
 pages = {5998--6008},
 title = {Attention is All you Need},
 url = {https://proceedings.neurips.cc/paper/2017/hash/3f5ee243547dee91fbd053c1c4a845aa-Abstract.html},
 year = {2017}
}

@inproceedings{Devlin2019BERTPO,
 address = {Minneapolis, Minnesota},
 author = {Devlin, Jacob  and
Chang, Ming-Wei  and
Lee, Kenton  and
Toutanova, Kristina},
 booktitle = {Proceedings of the 2019 Conference of the North {A}merican Chapter of the Association for Computational Linguistics: Human Language Technologies, Volume 1 (Long and Short Papers)},
 doi = {10.18653/v1/N19-1423},
 editor = {Burstein, Jill  and
Doran, Christy  and
Solorio, Thamar},
 pages = {4171--4186},
 title = {{BERT}: Pre-training of Deep Bidirectional Transformers for Language Understanding},
 url = {https://aclanthology.org/N19-1423},
 year = {2019}
}

@inproceedings{Radford2018ImprovingLU,
 author = {Alec Radford and Karthik Narasimhan},
 title = {Improving Language Understanding by Generative Pre-Training},
 url = {https://api.semanticscholar.org/CorpusID:49313245},
 year = {2018}
}

@inproceedings{Dosovitskiy2020AnII,
 author = {Alexey Dosovitskiy and
Lucas Beyer and
Alexander Kolesnikov and
Dirk Weissenborn and
Xiaohua Zhai and
Thomas Unterthiner and
Mostafa Dehghani and
Matthias Minderer and
Georg Heigold and
Sylvain Gelly and
Jakob Uszkoreit and
Neil Houlsby},
 bibsource = {dblp computer science bibliography, https://dblp.org},
 booktitle = {9th International Conference on Learning Representations, {ICLR} 2021,
Virtual Event, Austria, May 3-7, 2021},
 title = {An Image is Worth 16x16 Words: Transformers for Image Recognition
at Scale},
 url = {https://openreview.net/forum?id=YicbFdNTTy},
 year = {2021}
}

@inproceedings{Sun2017RevisitingUE,
 author = {Chen Sun and
Abhinav Shrivastava and
Saurabh Singh and
Abhinav Gupta},
 bibsource = {dblp computer science bibliography, https://dblp.org},
 booktitle = {International Conference on Computer Vision, {ICCV}, Venice,
Italy, October 22-29, 2017},
 doi = {10.1109/ICCV.2017.97},
 pages = {843--852},
 title = {Revisiting Unreasonable Effectiveness of Data in Deep Learning Era},
 url = {https://doi.org/10.1109/ICCV.2017.97},
 year = {2017}
}

@article{Papa2023ASO,
 author = {Lorenzo Papa and Paolo Russo and Irene Amerini and Luping Zhou},
 journal = {IEEE transactions on pattern analysis and machine intelligence},
 title = {A survey on efficient vision transformers: algorithms, techniques, and performance benchmarking},
 url = {https://api.semanticscholar.org/CorpusID:261531260},
 volume = {PP},
 year = {2023}
}

@inproceedings{Yang2020ResolutionAN,
 author = {Le Yang and
Yizeng Han and
Xi Chen and
Shiji Song and
Jifeng Dai and
Gao Huang},
 bibsource = {dblp computer science bibliography, https://dblp.org},
 booktitle = {2020 {IEEE/CVF} Conference on Computer Vision and Pattern Recognition,
{CVPR} 2020, Seattle, WA, USA, June 13-19, 2020},
 doi = {10.1109/CVPR42600.2020.00244},
 pages = {2366--2375},
 title = {Resolution Adaptive Networks for Efficient Inference},
 url = {https://doi.org/10.1109/CVPR42600.2020.00244},
 year = {2020}
}

@article{Lin2022SuperVT,
 author = {Mingbao Lin and Mengzhao Chen and Yu-xin Zhang and Ke Li and Yunhang Shen and Chunhua Shen and Rongrong Ji},
 journal = {International Journal of Computer Vision},
 pages = {3136-3151},
 title = {Super Vision Transformer},
 url = {https://api.semanticscholar.org/CorpusID:248986566},
 volume = {131},
 year = {2022}
}

@inproceedings{Chen2022CFViTAG,
 author = {Mengzhao Chen and
Mingbao Lin and
Ke Li and
Yunhang Shen and
Yongjian Wu and
Fei Chao and
Rongrong Ji},
 bibsource = {dblp computer science bibliography, https://dblp.org},
 booktitle = {Thirty-Seventh {AAAI} Conference on Artificial Intelligence, {AAAI}, Thirty-Fifth Conference on Innovative Applications of Artificial
Intelligence, {IAAI}, Thirteenth Symposium on Educational Advances
in Artificial Intelligence, {EAAI}},
 doi = {10.1609/AAAI.V37I6.25860},
 editor = {Brian Williams and
Yiling Chen and
Jennifer Neville},
 pages = {7042--7052},
 title = {CF-ViT: {A} General Coarse-to-Fine Method for Vision Transformer},
 url = {https://doi.org/10.1609/aaai.v37i6.25860},
 year = {2023}
}

@inproceedings{You2022CastlingViTCS,
 author = {Haoran You and
Yunyang Xiong and
Xiaoliang Dai and
Bichen Wu and
Peizhao Zhang and
Haoqi Fan and
Peter Vajda and
Yingyan Celine Lin},
 bibsource = {dblp computer science bibliography, https://dblp.org},
 booktitle = {Conference on Computer Vision and Pattern Recognition,
{CVPR}},
 doi = {10.1109/CVPR52729.2023.01387},
 pages = {14431--14442},
 title = {Castling-ViT: Compressing Self-Attention via Switching Towards Linear-Angular
Attention at Vision Transformer Inference},
 url = {https://doi.org/10.1109/CVPR52729.2023.01387},
 year = {2023}
}

@article{Wu2022TinyViTFP,
 author = {Kan Wu and Jinnian Zhang and Houwen Peng and Mengchen Liu and Bin Xiao and Jianlong Fu and Lu Yuan},
 journal = {ArXiv preprint},
 title = {TinyViT: Fast Pretraining Distillation for Small Vision Transformers},
 url = {https://arxiv.org/abs/2207.10666},
 volume = {abs/2207.10666},
 year = {2022}
}

@inproceedings{Liu2022NoisyQuantNB,
 author = {Yijiang Liu and
Huanrui Yang and
Zhen Dong and
Kurt Keutzer and
Li Du and
Shanghang Zhang},
 bibsource = {dblp computer science bibliography, https://dblp.org},
 booktitle = {Conference on Computer Vision and Pattern Recognition,
{CVPR}},
 doi = {10.1109/CVPR52729.2023.01946},
 pages = {20321--20330},
 title = {NoisyQuant: Noisy Bias-Enhanced Post-Training Activation Quantization
for Vision Transformers},
 url = {https://doi.org/10.1109/CVPR52729.2023.01946},
 year = {2023}
}

@inproceedings{Zhu2021VisionTP,
 author = {Mingjian Zhu and K. Han and Yehui Tang and Yunhe Wang},
 title = {Vision Transformer Pruning},
 url = {https://api.semanticscholar.org/CorpusID:233296620},
 year = {2021}
}

@article{Xu2023LGViTDE,
 author = {Guanyu Xu and Jiawei Hao and Li Shen and Han Hu and Yong Luo and Hui Lin and Ji Jia Shen},
 journal = {Proceedings of the 31st ACM International Conference on Multimedia},
 title = {LGViT: Dynamic Early Exiting for Accelerating Vision Transformer},
 url = {https://api.semanticscholar.org/CorpusID:260350893},
 year = {2023}
}

@article{Teerapittayanon2016BranchyNetFI,
 author = {Surat Teerapittayanon and Bradley McDanel and H. T. Kung},
 journal = {23rd International Conference on Pattern Recognition (ICPR)},
 pages = {2464-2469},
 title = {BranchyNet: Fast inference via early exiting from deep neural networks},
 url = {https://api.semanticscholar.org/CorpusID:2916466},
 year = {2016}
}

@inproceedings{deng2009imagenet,
 author = {Jia Deng and
Wei Dong and
Richard Socher and
Li{-}Jia Li and
Kai Li and
Fei{-}Fei Li},
 bibsource = {dblp computer science bibliography, https://dblp.org},
 booktitle = {Computer Society Conference on Computer Vision and Pattern
Recognition {(CVPR}},
 doi = {10.1109/CVPR.2009.5206848},
 pages = {248--255},
 title = {ImageNet: {A} large-scale hierarchical image database},
 url = {https://doi.org/10.1109/CVPR.2009.5206848},
 year = {2009}
}

@inproceedings{Le2015TinyIV,
 author = {Ya Le and Xuan S. Yang},
 title = {Tiny ImageNet Visual Recognition Challenge},
 url = {https://api.semanticscholar.org/CorpusID:16664790},
 year = {2015}
}

@inproceedings{Krizhevsky2009LearningML,
 author = {Alex Krizhevsky},
 title = {Learning Multiple Layers of Features from Tiny Images},
 url = {https://api.semanticscholar.org/CorpusID:18268744},
 year = {2009}
}

@misc{yoshioka2024visiontransformers,
 author = {Kentaro Yoshioka},
 howpublished = {\url{https://github.com/kentaroy47/vision-transformers-cifar10}},
 title = {vision-transformers-cifar10: Training Vision Transformers (ViT) and related models on CIFAR-10},
 year = {2024}
}

@inproceedings{Jiang2021AllTM,
 author = {Zihang Jiang and
Qibin Hou and
Li Yuan and
Daquan Zhou and
Yujun Shi and
Xiaojie Jin and
Anran Wang and
Jiashi Feng},
 bibsource = {dblp computer science bibliography, https://dblp.org},
 booktitle = {Advances in Neural Information Processing Systems 34: Annual Conference
on Neural Information Processing Systems, NeurIPS},
 editor = {Marc'Aurelio Ranzato and
Alina Beygelzimer and
Yann N. Dauphin and
Percy Liang and
Jennifer Wortman Vaughan},
 pages = {18590--18602},
 title = {All Tokens Matter: Token Labeling for Training Better Vision Transformers},
 url = {https://proceedings.neurips.cc/paper/2021/hash/9a49a25d845a483fae4be7e341368e36-Abstract.html},
 year = {2021}
}

@inproceedings{Liu2021SwinTH,
 author = {Ze Liu and
Yutong Lin and
Yue Cao and
Han Hu and
Yixuan Wei and
Zheng Zhang and
Stephen Lin and
Baining Guo},
 bibsource = {dblp computer science bibliography, https://dblp.org},
 booktitle = { International Conference on Computer Vision, {ICCV}},
 doi = {10.1109/ICCV48922.2021.00986},
 pages = {9992--10002},
 title = {Swin Transformer: Hierarchical Vision Transformer using Shifted Windows},
 url = {https://doi.org/10.1109/ICCV48922.2021.00986},
 year = {2021}
}

@inproceedings{Wang2021PyramidVT,
 author = {Wenhai Wang and
Enze Xie and
Xiang Li and
Deng{-}Ping Fan and
Kaitao Song and
Ding Liang and
Tong Lu and
Ping Luo and
Ling Shao},
 bibsource = {dblp computer science bibliography, https://dblp.org},
 booktitle = { International Conference on Computer Vision, {ICCV}},
 doi = {10.1109/ICCV48922.2021.00061},
 pages = {548--558},
 title = {Pyramid Vision Transformer: {A} Versatile Backbone for Dense Prediction
without Convolutions},
 url = {https://doi.org/10.1109/ICCV48922.2021.00061},
 year = {2021}
}

@inproceedings{Zheng2020RethinkingSS,
 author = {Sixiao Zheng and
Jiachen Lu and
Hengshuang Zhao and
Xiatian Zhu and
Zekun Luo and
Yabiao Wang and
Yanwei Fu and
Jianfeng Feng and
Tao Xiang and
Philip H. S. Torr and
Li Zhang},
 bibsource = {dblp computer science bibliography, https://dblp.org},
 booktitle = {Conference on Computer Vision and Pattern Recognition, {CVPR}},
 doi = {10.1109/CVPR46437.2021.00681},
 pages = {6881--6890},
 title = {Rethinking Semantic Segmentation From a Sequence-to-Sequence Perspective
With Transformers},
 url = {https://openaccess.thecvf.com/content/CVPR2021/html/Zheng\_Rethinking\_Semantic\_Segmentation\_From\_a\_Sequence-to-Sequence\_Perspective\_With\_Transformers\_CVPR\_2021\_paper.html},
 year = {2021}
}

@inproceedings{Fang2021MSGTransformerEL,
 author = {Jiemin Fang and
Lingxi Xie and
Xinggang Wang and
Xiaopeng Zhang and
Wenyu Liu and
Qi Tian},
 bibsource = {dblp computer science bibliography, https://dblp.org},
 booktitle = {Conference on Computer Vision and Pattern Recognition,{CVPR}},
 doi = {10.1109/CVPR52688.2022.01175},
 pages = {12053--12062},
 title = {MSG-Transformer: Exchanging Local Spatial Information by Manipulating
Messenger Tokens},
 url = {https://doi.org/10.1109/CVPR52688.2022.01175},
 year = {2022}
}

@article{Han2021DynamicNN,
 author = {Yizeng Han and Gao Huang and Shiji Song and Le Yang and Honghui Wang and Yulin Wang},
 journal = {IEEE Transactions on Pattern Analysis and Machine Intelligence},
 pages = {7436-7456},
 title = {Dynamic Neural Networks: A Survey},
 url = {https://api.semanticscholar.org/CorpusID:231855426},
 volume = {44},
 year = {2021}
}

@article{Carion2020EndtoEndOD,
 author = {Nicolas Carion and Francisco Massa and Gabriel Synnaeve and Nicolas Usunier and Alexander Kirillov and Sergey Zagoruyko},
 journal = {ArXiv preprint},
 title = {End-to-End Object Detection with Transformers},
 url = {https://arxiv.org/abs/2005.12872},
 volume = {abs/2005.12872},
 year = {2020}
}

@article{Thisanke2023SemanticSU,
 author = {Hans Thisanke and Chamli Deshan and Kavindu Chamith and Sachith Seneviratne and Rajith Vidanaarachchi and Damayanthi Herath},
 journal = {ArXiv preprint},
 title = {Semantic Segmentation using Vision Transformers: A survey},
 url = {https://arxiv.org/abs/2305.03273},
 volume = {abs/2305.03273},
 year = {2023}
}

@inproceedings{Arnab2021ViViTAV,
 author = {Anurag Arnab and
Mostafa Dehghani and
Georg Heigold and
Chen Sun and
Mario Lucic and
Cordelia Schmid},
 bibsource = {dblp computer science bibliography, https://dblp.org},
 booktitle = {International Conference on Computer Vision, {ICCV}},
 doi = {10.1109/ICCV48922.2021.00676},
 pages = {6816--6826},
 title = {ViViT: {A} Video Vision Transformer},
 url = {https://doi.org/10.1109/ICCV48922.2021.00676},
 year = {2021}
}

@inproceedings{Loshchilov2017DecoupledWD,
 author = {Ilya Loshchilov and
Frank Hutter},
 bibsource = {dblp computer science bibliography, https://dblp.org},
 booktitle = {7th International Conference on Learning Representations, {ICLR}},
 title = {Decoupled Weight Decay Regularization},
 url = {https://openreview.net/forum?id=Bkg6RiCqY7},
 year = {2019}
}

@inproceedings{Loshchilov2016SGDRSG,
 author = {Ilya Loshchilov and
Frank Hutter},
 bibsource = {dblp computer science bibliography, https://dblp.org},
 booktitle = {5th International Conference on Learning Representations, {ICLR}, Conference Track Proceedings},
 title = {{SGDR:} Stochastic Gradient Descent with Warm Restarts},
 url = {https://openreview.net/forum?id=Skq89Scxx},
 year = {2017}
}

@inproceedings{Ionescu2016HowHC,
 author = {Radu Tudor Ionescu and
Bogdan Alexe and
Marius Leordeanu and
Marius Popescu and
Dim P. Papadopoulos and
Vittorio Ferrari},
 bibsource = {dblp computer science bibliography, https://dblp.org},
 booktitle = {Conference on Computer Vision and Pattern Recognition,
{CVPR}},
 doi = {10.1109/CVPR.2016.237},
 pages = {2157--2166},
 title = {How Hard Can It Be? Estimating the Difficulty of Visual Search in
an Image},
 url = {https://doi.org/10.1109/CVPR.2016.237},
 year = {2016}
}

@inproceedings{Li2016PruningFF,
 author = {Hao Li and
Asim Kadav and
Igor Durdanovic and
Hanan Samet and
Hans Peter Graf},
 bibsource = {dblp computer science bibliography, https://dblp.org},
 booktitle = {5th International Conference on Learning Representations, {ICLR}, Conference Track Proceedings},
 title = {Pruning Filters for Efficient ConvNets},
 url = {https://openreview.net/forum?id=rJqFGTslg},
 year = {2017}
}

@inproceedings{Sandler2018MobileNetV2IR,
 author = {Mark Sandler and
Andrew G. Howard and
Menglong Zhu and
Andrey Zhmoginov and
Liang{-}Chieh Chen},
 bibsource = {dblp computer science bibliography, https://dblp.org},
 booktitle = {Conference on Computer Vision and Pattern Recognition,
{CVPR}},
 doi = {10.1109/CVPR.2018.00474},
 pages = {4510--4520},
 title = {MobileNetV2: Inverted Residuals and Linear Bottlenecks},
 url = {http://openaccess.thecvf.com/content\_cvpr\_2018/html/Sandler\_MobileNetV2\_Inverted\_Residuals\_CVPR\_2018\_paper.html},
 year = {2018}
}

@article{Hinton2015DistillingTK,
 author = {Geoffrey E. Hinton and Oriol Vinyals and Jeffrey Dean},
 journal = {ArXiv preprint},
 title = {Distilling the Knowledge in a Neural Network},
 url = {https://arxiv.org/abs/1503.02531},
 volume = {abs/1503.02531},
 year = {2015}
}

@inproceedings{Hubara2016BinarizedNN,
 author = {Itay Hubara and
Matthieu Courbariaux and
Daniel Soudry and
Ran El{-}Yaniv and
Yoshua Bengio},
 bibsource = {dblp computer science bibliography, https://dblp.org},
 booktitle = {Advances in Neural Information Processing Systems 29: Annual Conference
on Neural Information Processing Systems},
 editor = {Daniel D. Lee and
Masashi Sugiyama and
Ulrike von Luxburg and
Isabelle Guyon and
Roman Garnett},
 pages = {4107--4115},
 title = {Binarized Neural Networks},
 url = {https://proceedings.neurips.cc/paper/2016/hash/d8330f857a17c53d217014ee776bfd50-Abstract.html},
 year = {2016}
}

@article{Krishnamoorthi2018QuantizingDC,
 author = {Raghuraman Krishnamoorthi},
 journal = {ArXiv preprint},
 title = {Quantizing deep convolutional networks for efficient inference: A whitepaper},
 url = {https://arxiv.org/abs/1806.08342},
 volume = {abs/1806.08342},
 year = {2018}
}

@article{Demir2024EarlyexitCN,
 author = {Edanur Demir and Emre Akbas},
 journal = {ArXiv preprint},
 title = {Early-exit Convolutional Neural Networks},
 url = {https://arxiv.org/abs/2409.05336},
 volume = {abs/2409.05336},
 year = {2024}
}

@inproceedings{Fan2021MultiscaleVT,
 author = {Haoqi Fan and
Bo Xiong and
Karttikeya Mangalam and
Yanghao Li and
Zhicheng Yan and
Jitendra Malik and
Christoph Feichtenhofer},
 bibsource = {dblp computer science bibliography, https://dblp.org},
 booktitle = {International Conference on Computer Vision, {ICCV}},
 doi = {10.1109/ICCV48922.2021.00675},
 pages = {6804--6815},
 title = {Multiscale Vision Transformers},
 url = {https://doi.org/10.1109/ICCV48922.2021.00675},
 year = {2021}
}

\vfill

\end{document}